\documentclass[11pt, oneside]{article}   	
\usepackage{geometry}                		
\usepackage{graphicx}				
\usepackage{amssymb}

\usepackage{amsmath}

\DeclareMathOperator*{\argmax}{arg\,max}

\title{Reinforcement Learning Agents in Colonel Blotto}
\author{Joseph Christian G. Noel}
\date{}							

\begin{document}
\maketitle

\begin{abstract}
Models and games are simplified representations of the world. There are many different kinds of models, all differing in complexity and which aspect of the world they allow us to further our understanding of. In this paper we focus on a specific instance of agent-based models, which uses reinforcement learning (RL) to train the agent how to act in its environment. Reinforcement learning agents are usually also Markov processes, which is another type of model that can be used. We test this reinforcement learning agent in a Colonel Blotto environment\footnote{Source code for the experiments can be found at https://github.com/jinonoel/BlottoRL}, and measure its performance against Random agents as its opponent. We find that the RL agent handily beats a single opponent, and still performs quite well when the number of opponents are increased. We also analyze the RL agent and look at what strategies it has arrived by looking at the actions that it has given the highest and lowest $Q$-values. Interestingly, the optimal strategy for playing multiple opponents is almost the complete opposite of the optimal strategy for playing a single opponent.

 \end{abstract}
 
\section{Colonel Blotto}

Colonel Blotto is a constant-sum game proposed by Émile Borel in 1921\cite{Borel1953}. In the game two or more players distribute resources (or coins) over several fronts in a battlefield. A player wins a front if they have allocated more resources to it than the other player. The objective of the game is to win more fronts than your opponent.

\hfill

A notable thing about Colonel Blotto is that there is no optimal best strategy that will beat all other possible strategies. This means that the best strategy can change in a given game depending on the opponent and what strategy the opponent is using. One way to play Colonel Blotto would be to update your strategy over time as you play the same opponent repeatedly. If the opponent's strategy is constant or doesn't change much, it must be possible to eventually "learn" a strategy that would win against this opponent more often than it loses. One way to do this would be to create an agent-based model that would update its strategy as it plays an opponent multiple times.

\section{Reinforcement Learning}

Reinforcement learning is a form of agent-based modeling. In RL an agent learns by performing actions which changes the state of an environment. After each action the agent may also receive a "reward" from the environment, whose value should depend on how close the agent is to what we want it to achieve. The goal for the agent is to maximize the cumulative reward that it receives over time. After a series of actions, the agent eventually reaches a goal state or terminal state, which signifies the end of an episode. The environment is then reset and the agent starts again from an initial state and the process repeats itself.

\hfill

Formally, an RL model is a set of states $S$, a set of actions $A$, and transition rules between states depending on the action taken. For state $s \in S$ at time $t$, an agent performs an action $a \in A$, moves to a new state $s'$ and receives a reward $r_t$.  The agent follows a policy $\pi$ for choosing which action to take, and $\pi(s,a)$ is the probability mapping of an agent taking action $a$ while in state $s$. The goal of the agent is to maximize the expected reward $Rt$,

\begin{equation}
R_t = \sum_{k=0}^{\infty}\lambda^kr_{t+k}
\end{equation}

where $0 \leq \gamma \leq 1$ is a discount factor for handling infinite horizons.

\subsection{Markov Decision Process}

Calculating the optimal policy in general is a hard problem, especially if we need to keep track of the history of all states, actions, and rewards. To simplify things, reinforcement learning problems are usually constructed as Markov models, such that the information from the history of all states, actions, and rewards before time $t$ is encapsulated in the current state at time $t$. This is the Markov property, and control optimization tasks which exhibit this property are called Markov decision processes (MDP). Formally, the Markov property states that for transition probability function $Pr$,

\begin{equation}
Pr(s_{t+1}, r_{t+1} | s_t, a_t) = Pr(s_{t+1}, r_{t+1} | s_t,a_t, r_t, s_{t-1}, a_{t-1}, r_{t-1},...,s_0, a_0, r_0)
\end{equation}

\subsection{Value Function}
The state-action value function $Q$ is the estimate of the expected rewards the agent will receive by being at state $s$ at time $t$, taking action $a$, and then following policy $\pi$ from $t+1$ onwards. 

\begin{equation}
Q^\pi(s,a) = \mathbb{E}_\pi(R_t | s_t, a_t)
\end{equation}

The reinforcement learning problem can be reduced to being able to calculate $Q(s,a)$ accurately for all $s \in S$ and  $a \in A$. The agent can then follow the optimal policy $\pi^*$ by simply choosing the action $a$ that maximizes $Q(s,a)$ for all $s \in S$.

\begin{equation}
\pi^*(s) = \argmax_a{Q(s,a)}
\end{equation}

\subsection{Q-Learning}

Q-Learning is one of the basic reinforcement learning methods, and takes full advantage of the Markov property of the RL model. It approximates the state-action value function $Q(s,a)$ in a recursive manner at time $t$ using a weighted average of the current value and the new information retrieved from applying the action $a_{t+1}$ and receiving subsequent reward $r_{t+1}$ from the environment. 

\hfill

At each time step Q-Learning approximates the optimal policy by iteratively updating $Q$ in the following manner, where $\alpha > 0$ is the learning rate, and $0 \leq \gamma \leq 1$ is again the discount factor:

\begin{equation}
Q(s_t, a_t) = Q(s_t, a_t) + \alpha[r_{t+1} + \gamma \max_aQ(s_{t+1}, a_{t+1}) - Q(s_t, a_t)]
\end{equation}

\section{Experiments}

We now show how a Q-Learning agent performs against a Random agent in Colonel Blotto.

\subsection{Experimental Setup}

We setup a 2-player Colonel Blotto game with 3 fields and 10 coins per player. This provides a total of 66 possible actions $a$ (ways to distribute the 10 coins on the 3 fields) that the players can choose from. The players aim to distribute their 10 coins among the 3 fields in such a way that their coins are greater than their opponent's coins in each field. The winner of the game is the player that wins more fields than their opponent, and receives a reward of $r=1$ from the environment. The loser of the game receives a reward of $r=-1$. A single game constitutes a single episode in reinforcement learning. We count how many games each player has won over time and show the results.

\hfill

Our reinforcement learning agent uses the Q-Learning algorithm for approximating the optimal policy. We use $\alpha=0.1$ as the learning rate and $\gamma=1$ as the discount factor. We also set $\epsilon=0.2$ as the exploration rate for the agent. For the opponent, we use a Random agent that will select one of the 66 possible actions at random with equal probability for all of them. The experiments were run using the OpenSpiel\cite{OpenSpiel} environment simulator.

\subsection{Results}

We run 1 million games of Colonel Blotto to see the performance of the reinforcement learning agent and show the results in Figure 1. While the RL agent eventually wins significantly more games, it takes awhile before the RL agent really begins to start defeating the Random agent regularly. In fact, if we look at the just the first 200 games (Figure 2), both agents win games at practically similar rates to start. However after 200 games the better win rate of the RL agent can already be seen.

\begin{figure}[h]
    \centering
    \includegraphics[scale=0.5]{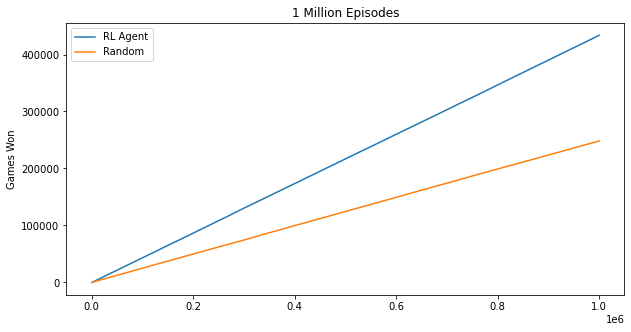}
    \caption{Performance of RL Agent vs Random Agent}
    \label{fig:mesh1}
\end{figure}

\begin{figure}[h]
    \centering
    \includegraphics[scale=0.5]{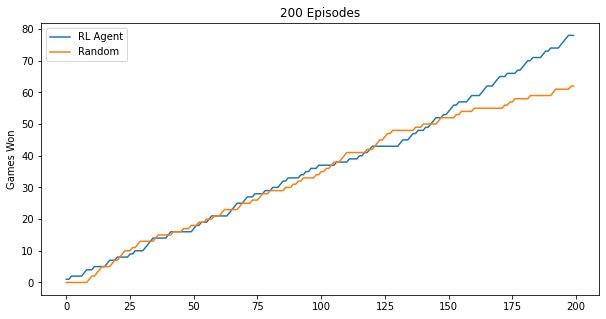}
    \caption{First 200 games of RL Agent vs Random Agent}
    \label{fig:mesh1}
\end{figure}

We also dive deeper and look at the $Q$-values that the agent has assigned to each of the 66 possible actions, to better understand what strategy it has arrived at. Below we show the top 4 and the bottom 4 actions based on their $Q$-value scores. The 3 numbers for each action signify the number of coins that the RL agent assigns for each of the 3 fields.

\hfill

\begin{center}
\begin{tabular}{ |c| } 
 \hline
 \textbf{Top 4 Actions} \\ 
 \hline
 [4, 0, 6] \\ 
 \hline
 [5, 0, 5] \\ 
 \hline
 [0, 4, 6] \\
 \hline
 [6, 1, 3] \\

 \hline
\end{tabular}
\quad
\begin{tabular}{ |c| } 
 \hline
\textbf{Worst 4 Actions} \\ 
 \hline
 [10, 0, 0]\\ 
 \hline
[0, 10, 0]\\
 \hline
[0, 8, 2]\\
 \hline
[0, 0, 10] \\
 \hline
\end{tabular}
\end{center}

\hfill

As can be seen, the actions with the lowest $Q$-values are the ones where the agent puts all their coins in just one front, guaranteeing a loss when their opponent easily wins the other two fronts. On the opposite end, the best actions are the ones where the agent splits their coins almost evenly among just two fronts, with 0 coins allocated to the third front. By sacrificing one of the fronts, they give themselves the best chance to win the other two fronts they focused on, and have a high probability of winning the game in the process. Against a Random agent, this results in the high win rates that were shown in the earlier graphs.

\subsection{Games with 3 or More Players}

Next we analyze what happens when we extend Colonel Blotto to games of 3 or more players. Colonel Blotto was originally envisioned as a 2-player game, and when extending to more than 2 players we need to properly define first what will be the winning conditions for the game and how that win is divided in case there are multiple winners. For this paper we take the definition of the implementation of Colonel Blotto in OpenSpiel\footnote{Implementation code can be found in https://github.com/deepmind/open\_spiel}:

\begin{enumerate}
  \item Iterate over every front and get the players that have allocated the most coins for each one. If there is a single winner for a front, give that player 1 point. However if there are multiple winners for a front, don't assign any points to those players.
  \item Find the players with the most numbers of points. If only one player has the highest points count then there is only a single winner for the game. Otherwise the players with the same highest points count are all considered the winners.
  \item Each of the winners gets a score of $\frac{1}{num\_winners}$    
\end{enumerate}

The returned score is also used as the environment reward for the winners in reinforcement learning. For the losers of the game, they get a reward of $\frac{-1}{num\_losers}$.

\hfill

Using this method for calculating winning games in Colonel Blotto of games with 3 or more players, we play 1 million games while increasing the number of Random agent opponents from 1 to 10 and show the results at each opponent count. Figure 3 shows that there is a clear drop in games won for the RL agent when the number of opponents is increased from 1 to 2. The wins then increase when the number of opponents is set to 3, and then stays in that range without much variation as the number of opponents keep increasing. 

\hfill

\begin{figure}[h]
    \centering
    \includegraphics[scale=0.5]{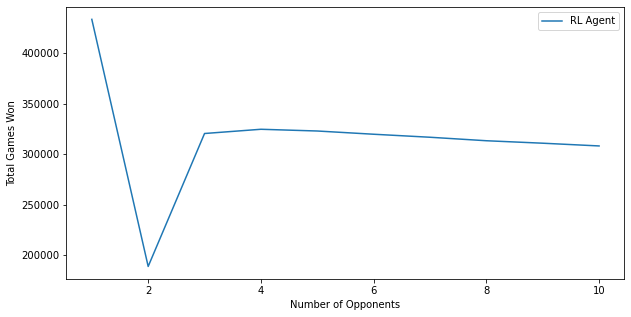}
    \caption{Won games by RL agent again different number of opponents}
    \label{fig:mesh1}
\end{figure}

\begin{figure}[h]
    \centering
    \includegraphics[scale=0.5]{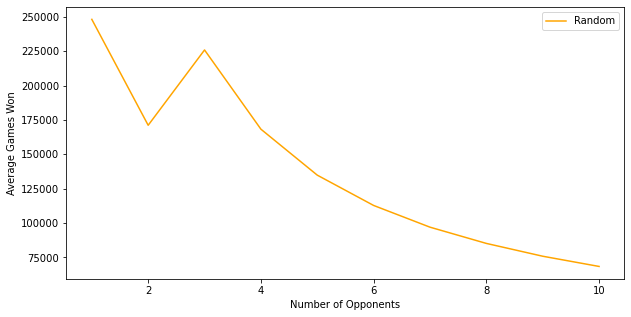}
    \caption{Average wins of a single Random opponent}
    \label{fig:mesh1}
\end{figure}

It may seem that the RL agent isn't doing well when more opponents are added into the game. However one thing to remember is that although the number of wins by the RL agent seems to be almost constant, the number of the opponents increasing mean that the Random agents are getting fewer wins each. In Figure 4 we show the average count of won games of each Random agent as their number increases. Since all the opponents share the same Random strategy, the variance between each of them is very small and the average count shown here are representative for all of them.

\hfill

Finally we again look at the $Q$-values and see which actions the RL agent has deemed best. For the table below we show the actions with the 4 highest and 4 lowest $Q$-values against 10 opponents.

\hfill 

\begin{center}
\begin{tabular}{ |c| } 
 \hline
 \textbf{Top 4 Actions} \\ 
 \hline
[10, 0, 0]] \\ 
 \hline
 [5, 0, 5] \\ 
 \hline
[9, 1, 0]\\
 \hline
[0, 10, 0]\\
 \hline
\end{tabular}
\quad
\begin{tabular}{ |c| } 
 \hline
\textbf{Worst 4 Actions} \\ 
 \hline
[4, 4, 2]\\ 
 \hline
[0, 5, 5] \\
 \hline
 [3, 3, 4]\\
 \hline
 [2, 5, 3]\\

 \hline
\end{tabular}

\end{center}

We find that the previous worst strategies against 1 opponent, where the agent sacrifices two fronts, are now the best strategies when playing against 10 opponents. The strategy for the RL agent now is to ensure that it wins one of the fronts against all opponents. This is mainly due to the winning condition detailed above where a player needs to win at least one front against all opponents to have a chance at winning a game. This also shows how important it is to properly set the rewards being returned by the environment in reinforcement learning. By changing the conditions of the rewards in such a way when we moved to 3 or more players, we have trained our RL agent to have very different strategies compared to earlier.

\section{Conclusion}

We have shown how an RL agent using Q-Learning performs against a Random agent in 1 million games of Colonel Blotto. It takes almost 200 episodes for the RL agent to start differentiating itself to the Random agent, and after 1 million episodes the RL agent has clearly shown its dominance.

\hfill

We also investigated what strategies the RL agent has learned through its repeated playing, and find that its preferred strategy is to sacrifice one of the fronts and focus on the other 2 fronts, splitting the coins almost evenly between them. This was shown empirically to be a a winning strategy against a single Random agent.

\hfill

Lastly, we also investigated what happens when we increase the number of opponents from 1 to 10. We found that the number of games the RL agent wins is significantly less when playing 2 or more opponents. However, the RL agent still performs much better and wins more games compared to each single Random agent. We also found that the optimal strategy when playing 10 opponents is the complete opposite to playing just 1 opponent, mainly due to how the winning conditions are designed and how rewards are being returned from the environment.

\nocite{*}
\bibliographystyle{abbrv}
\bibliography{References}

\begin{thebibliography}{1}

\bibitem{Borel1953}
Emile Borel.
\newblock The theory of play and integral equations with skew symmetric
  kernels.
\newblock 1953.

\bibitem{OpenSpiel}
Marc Lanctot, Edward Lockhart, Jean-Baptiste Lespiau, Vinicius Zambaldi,
  Satyaki Upadhyay, Julien P\'{e}rolat, Sriram Srinivasan, Finbarr Timbers,
  Karl Tuyls, Shayegan Omidshafiei, Daniel Hennes, Dustin Morrill, Paul Muller,
  Timo Ewalds, Ryan Faulkner, J\'{a}nos Kram\'{a}r, Bart~De Vylder, Brennan
  Saeta, James Bradbury, David Ding, Sebastian Borgeaud, Matthew Lai, Julian
  Schrittwieser, Thomas Anthony, Edward Hughes, Ivo Danihelka, and Jonah
  Ryan-Davis.
\newblock {OpenSpiel}: A framework for reinforcement learning in games.
\newblock {\em CoRR}, abs/1908.09453, 2019.

\bibitem{Sutton1998}
Richard~S. Sutton and Andrew~G. Barto.
\newblock {\em Reinforcement Learning: An Introduction}.
\newblock The MIT Press, second edition, 2018.

\end{thebibliography}

\end{document}